\documentclass[10pt,twocolumn,letterpaper]{article}

\usepackage{cvpr}              % To produce the CAMERA-READY version
\definecolor{cvprblue}{rgb}{0.21,0.49,0.74}
\usepackage[pagebackref,breaklinks,colorlinks,allcolors=cvprblue]{hyperref}

\def\paperID{8035} % *** Enter the Paper ID here
\def\confName{CVPR}
\def\confYear{2026}

\usepackage{multirow}

\title{\includegraphics[width=0.43in]{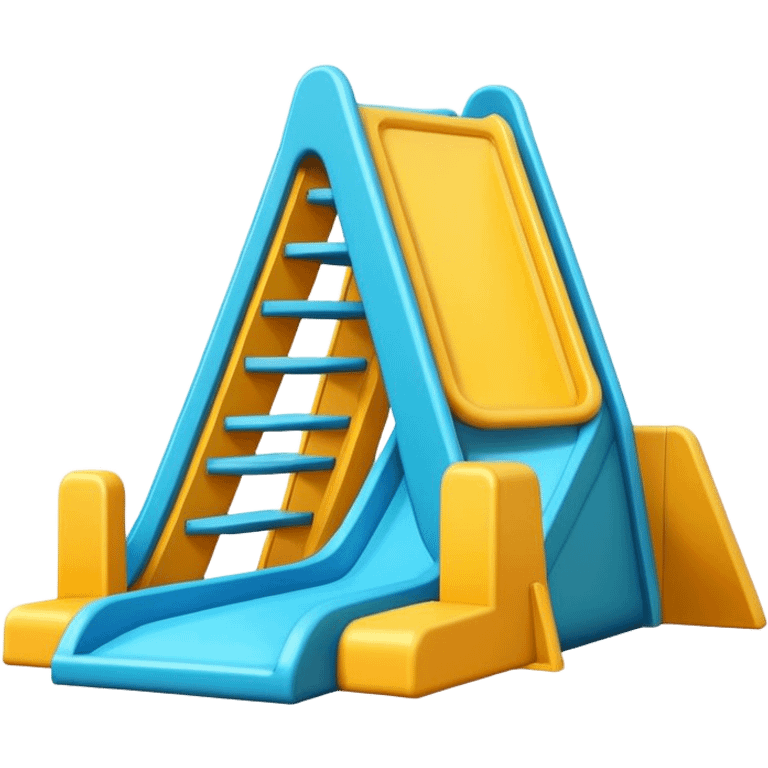}~VideoScaffold: Elastic-Scale Visual Hierarchies for Streaming Video Understanding in MLLMs}

\author{
{Naishan Zheng} \textsuperscript{1}\thanks{Work done during internship at Ant Group.}, 
{Jie Huang} \textsuperscript{1},  
{Qingpei Guo} \textsuperscript{2},  
{Feng Zhao} \textsuperscript{1}\thanks{Corresponding author.}, \\
\textsuperscript{1}University of Science and Technology of China
\hspace{0.2cm}\textsuperscript{2}Ant Group
}

\begin{document}

\maketitle
\begin{abstract}
% \footnotetext[1]{Work done during internship at Ant Group.}
% \footnotetext[2]{Corresponding author.}
Understanding long videos with multimodal large language models (MLLMs) remains challenging due to the heavy redundancy across frames and the need for temporally coherent representations. Existing static strategies—such as sparse sampling, frame compression, and clustering—are optimized for offline settings and often produce fragmented or over-compressed outputs when applied to continuous video streams.
We present VideoScaffold, a dynamic representation framework designed for streaming video understanding. It adaptively adjusts event granularity according to video duration while preserving fine-grained visual semantics.
VideoScaffold introduces two key components: Elastic-Scale Event Segmentation (EES), which performs prediction-guided segmentation to dynamically refine event boundaries, and Hierarchical Event Consolidation (HEC), which progressively aggregates semantically related segments into multi-level abstractions. 
Working in concert, EES and HEC enable VideoScaffold to transition smoothly from fine-grained frame understanding to abstract event reasoning as the video stream unfolds.
Extensive experiments across both offline and streaming video understanding benchmarks demonstrate that VideoScaffold achieves state-of-the-art performance. The framework is modular and plug-and-play, seamlessly extending existing image-based MLLMs to continuous video comprehension.
The code is available at https://github.com/zheng980629/VideoScaffold.
\end{abstract}    
\section{Introduction}
\label{sec:intro}
Large Language Models (LLMs)~\cite{yang2024qwen2,cai2024internlm2,grattafiori2024llama,achiam2023gpt} have demonstrated remarkable reasoning and comprehension abilities across a wide range of linguistic tasks. Building on this success, their extension to the visual domain has led to a surge of image-centric multimodal models~\cite{llava,llava_onevision,internvl3}, achieving impressive progress in open-world visual understanding.
However, scaling these capabilities to videos introduces unique challenges. Unlike static images, videos exhibit long, redundant temporal sequences and complex causal dependencies, causing both computational inefficiency and semantic dilution across frames.

\begin{figure*}[!t]
\centering
\includegraphics[width=0.98\textwidth]{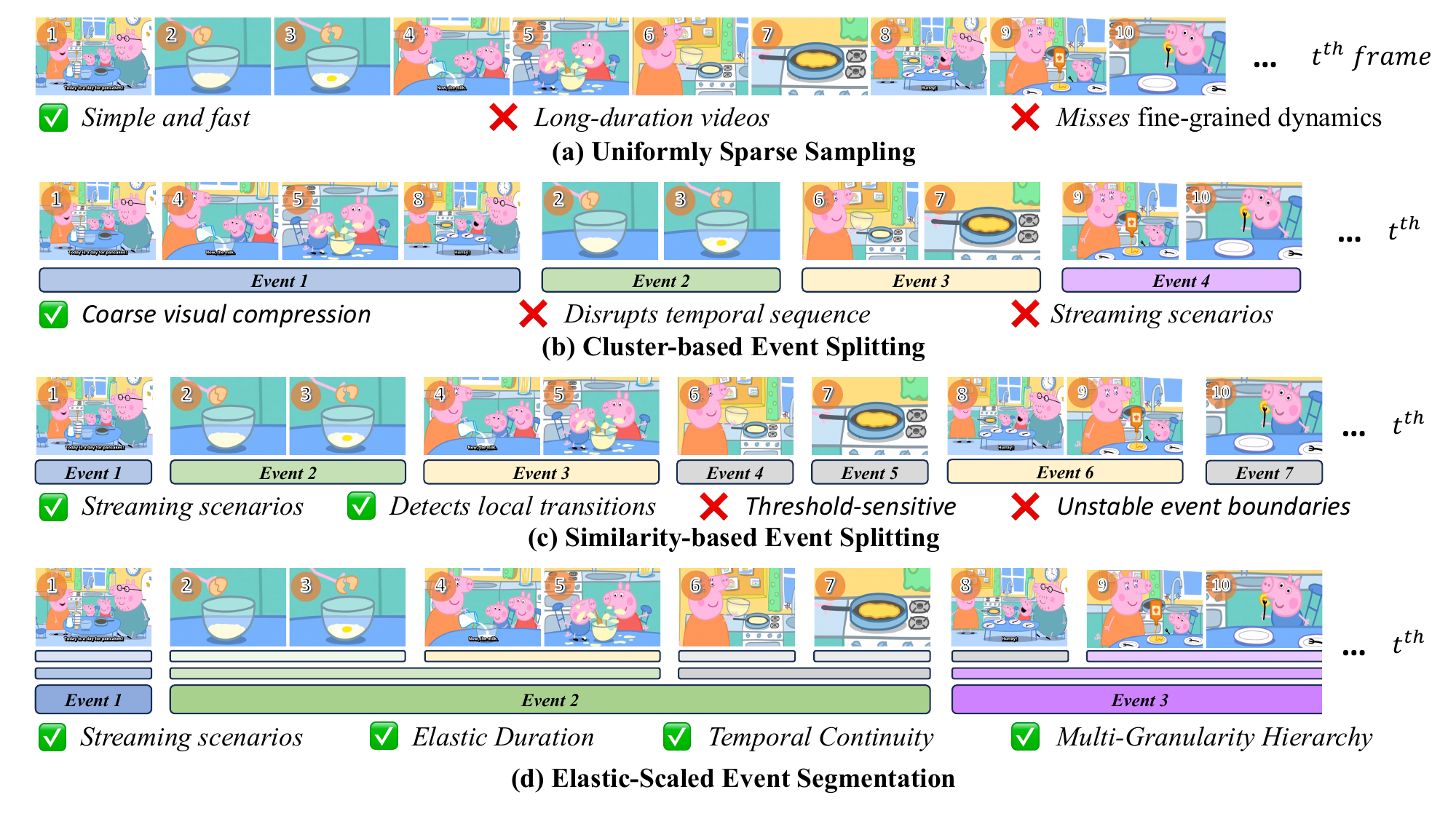}
\vspace{-0.2cm}
\caption{Comparative analysis of frame-sampling strategies for video understanding.
(a) Uniform sampling is common in offline settings but lacks adaptability to dynamic content.figs/
(b) Cluster-based segmentation groups visually similar frames into events while ignoring temporal order, producing disordered segments.
(c) Similarity-based segmentation detects event changes from adjacent-frame similarity, yet fixed thresholds often cause over-fragmentation or incorrect merging.
(d) Our Elastic-scale Event Segmentation (EES) continuously refines boundaries and preserves temporal continuity, forming elastic multi-granularity structures that adapt to varying video durations.}
\label{fig:elastic_event}
\vspace{-0.2cm}
\end{figure*}

A common approach is to sample a fixed number of frames~\cite{video_llava} and apply spatiotemporal pooling~\cite{llava_onevision,pllava,llava_interleave} to reduce visual tokens. However, such static compression often discards critical details, especially in long videos. To address this, LLaMA-VID~\cite{llama_vid} compresses each frame into two tokens, a text-guided context token and a visual content token, improving efficiency but still relying on uniform frame processing. Recent advances in event-based segmentation~\cite{chat_univi,dynfocus,qian2024streaming} mitigate temporal redundancy by grouping visually similar frames into coherent events. Chat-UniVi~\cite{chat_univi} employs multi-scale clustering, while FastVID~\cite{shen2025fastvid} detects boundaries via frame similarity. Although effective in offline scenarios. Although effective in offline scenarios, these methods encounter challenges in streaming video understanding, where models must reason causally without full video access.

\emph{In streaming settings, frames arrive continuously, the overall sequence length is unknown, and future events are inherently unpredictable.}
Under the causal constraint, where only past and current frames are observable, both clustering and threshold-based approaches demonstrate notable limitations.
As shown in Fig.~\ref{fig:elastic_event}(b), Clustering-based methods rely on global grouping across the entire sequence and thereby disrupt temporal order, violating the causal dependencies that govern event progression.
In contrast, threshold-based segmentation (Fig.~\ref{fig:elastic_event}(c)) operates solely on local similarity and lacks cross-temporal perception, often resulting in excessively fragmented and temporally incoherent event clips.
Such segmentation either over-divides continuous scenes or over-compresses semantically diverse content, ultimately impairing temporal coherence and semantic fidelity.
These limitations motivate the development of a streaming-compatible segmentation and compression framework that preserves temporal continuity while generating compact and semantically consistent representations across varied video durations.

In this paper, as illustrated in Fig.~\ref{fig:framework}, we present VideoScaffold, a unified framework for dynamic event segmentation and consolidation that preserves the temporal continuity characteristic of streaming video.
To effectively handle videos of varying lengths, we propose an \textbf{E}lastic-scale \textbf{E}vent \textbf{S}egmentation (\textbf{EES}) strategy.
Drawing inspiration from the human perceptual system, where abrupt changes in visual stimuli trigger attention and coherent content is summarized through memory, EES employs a \emph{next-frame prediction} mechanism to perform segmentation in a zero-shot, streaming-compatible manner. 
The prediction results are used to dynamically refine event boundaries, either initiating a new event when a significant deviation occurs or merging frames into an existing one when continuity is maintained.
By quantifying prediction errors over time, as shown in Fig.~\ref{fig:event_segment}, EES incrementally evolves event representations, forming a multi-granularity hierarchy that transitions from fine-grained to abstract levels as the video unfolds.
Through the integration of prediction-guided segmentation and dynamic structural evolution, EES constructs an elastic event hierarchy that faithfully aligns with the temporal continuity and evolving dynamics of streaming video.
Upon segmentation, we introduce a \textbf{H}ierarchical \textbf{E}vent \textbf{C}onsolidation (\textbf{HEC}) module to aggregate visual embeddings within each event structure.
Leveraging the prediction dynamics and multi-granularity hierarchy produced by EES, as demonstrated in Sec.~\ref{sec:abaltion}, HEC identifies essential frames as semantic anchors and performs bottom-up aggregation to construct progressively abstract event-level representations.
This hierarchical consolidation captures both local visual detail and global semantic context, enabling scalable and temporally coherent video understanding.

Our main contributions are following:
\begin{itemize}
    \item We propose VideoScaffold, an elastic event representation framework for streaming video understanding that adaptively scales event granularity with varying durations while preserving temporal and semantic coherence.
    \item We introduce a prediction-guided segmentation and hierarchical representation mechanism enabling dynamic event abstraction under streaming constraints, implemented via the proposed Elastic-scale Event Segmentation (EES) and Hierarchical Event Consolidation (HEC).
    \item VideoScaffold achieves state-of-the-art performance on both offline and streaming benchmarks and can be seamlessly integrated with existing image-based MLLMs for continuous video comprehension.
\end{itemize}
\section{Related Work}
\label{sec:related_work}
\textbf{Multi-Modal Large Language Model.}
Large Language Models (LLMs)~\cite{devlin2019bert,raffel2020exploring,yang2024qwen2,cai2024internlm2,grattafiori2024llama,achiam2023gpt} have advanced language comprehension, reasoning, and problem solving.
To extend these capabilities to vision, recent studies~\cite{blip2,llava,minigpt,flamingo,qwen2_vl,internvl,mplug_o1} align visual encoders with LLMs through lightweight adaptation, enabling open-ended image understanding and cross-modal reasoning.

Video understanding poses unique challenges due to the high redundancy of visual information across temporal sequences, making direct adaptation of image-based bridging methods inefficient.
To mitigate this redundancy, recent works adopt spatial pooling~\cite{llava_onevision,llava}, spatio-temporal pooling~\cite{pllava,liu2024kangaroo,luo2023valley}, and Q-Former-style adapters~\cite{videochat,video_llama,mplug_o2} to compress repetitive tokens and enhance temporal modeling.
In parallel, memory-bank mechanisms~\cite{ryoo2023token,wang2024videollamb,flash_vstream,ma_llm,song2024moviechat,song2024moviechat+} store representative visual embeddings for long videos.
For instance, MA-LLM~\cite{ma_llm} maintains both raw and abstracted memory banks to compress redundant information, while MovieChat~\cite{song2024moviechat} introduces a hierarchical memory system that distinguishes short- and long-term contexts, reducing storage cost while preserving key temporal cues.
Another line of work leverages text-driven selection~\cite{llama_vid,liu2025hybrid,liang2024end,videoespresso,wang2023vaquita} to retrieve salient frames from redundant video streams.
LLaMA-VID~\cite{llama_vid} represents each frame with a text-guided token and a content token for efficient long-video encoding, and VideoEspresso~\cite{videoespresso} dynamically selects question-relevant frames to enhance query-specific reasoning.
Finally, clustering-based methods such as Chat-UniVi~\cite{chat_univi} and DynFocus~\cite{dynfocus} reduce redundancy by grouping visually similar frames into events and encoding them with varying granularity.
However, these offline approaches rely on full temporal context, making them unsuitable for streaming scenarios with uncertain video durations and causal processing, where unpredictable token lengths disrupt stable temporal modeling.

\textbf{Streaming Video Understanding.}
Current video MLLMs primarily operate in an offline paradigm, requiring complete video access before prediction.
Such methods are unsuitable for streaming scenarios, where queries may emerge at arbitrary moments during playback.
To overcome this limitation, several streaming-oriented approaches have recently been proposed.
Flash-VStream~\cite{flash_vstream} dynamically maintains multi-granular memory banks—spatial, temporal, and abstract—during frame-by-frame streaming.
VideoLLM-Online~\cite{chen2024videollm} reformulates temporal understanding as a streaming dialogue, allowing the model to decide when to remain silent or respond based on contextual cues.
VideoLLaMB~\cite{wang2024videollamb} and VideoStream~\cite{qian2024streaming} propagate memory to distill essential historical cues into compact embeddings for efficient retrieval.
However, most existing solutions rely on fixed-length frame chunks, resulting in fragmented event boundaries, redundant representations, or over-compressed summaries.
In contrast, our method introduces a streaming elastic-scale segmentation strategy that adaptively adjusts event granularity to video duration, ensuring both temporal coherence and efficient representation for streaming video understanding.
\section{Methodology}
\subsection{Problem Formulation}
A streaming video can be represented as 
$V = \{v_1, v_2, \ldots, v_T\}$
where each $v_t \in \mathbb{R}^{N \times d}$ denotes the feature embedding of the $t$-th frame extracted by a vision encoder, $N$ is the number of image patches, and $d$ is the feature dimension. 
At time step $t$, the model can only access the observed segment 
$V_{[1,t]} = \{v_1, v_2, \ldots, v_t\}$
without access to future frames.
Given an instruction $C$, MLLM produces a response 
$R = f\big(C,\, g(V_{[1,t]})\big)$. 
$g(\cdot)$ is a visual representation function that encodes the observed stream into a compact, semantically coherent representation. 
This setting imposes a \textbf{causal constraint}, requiring the model to reason only over past and current content—unlike offline video understanding, which assumes full-sequence access.

\begin{figure*}[!ht]
\centering
\includegraphics[width=0.98\textwidth]{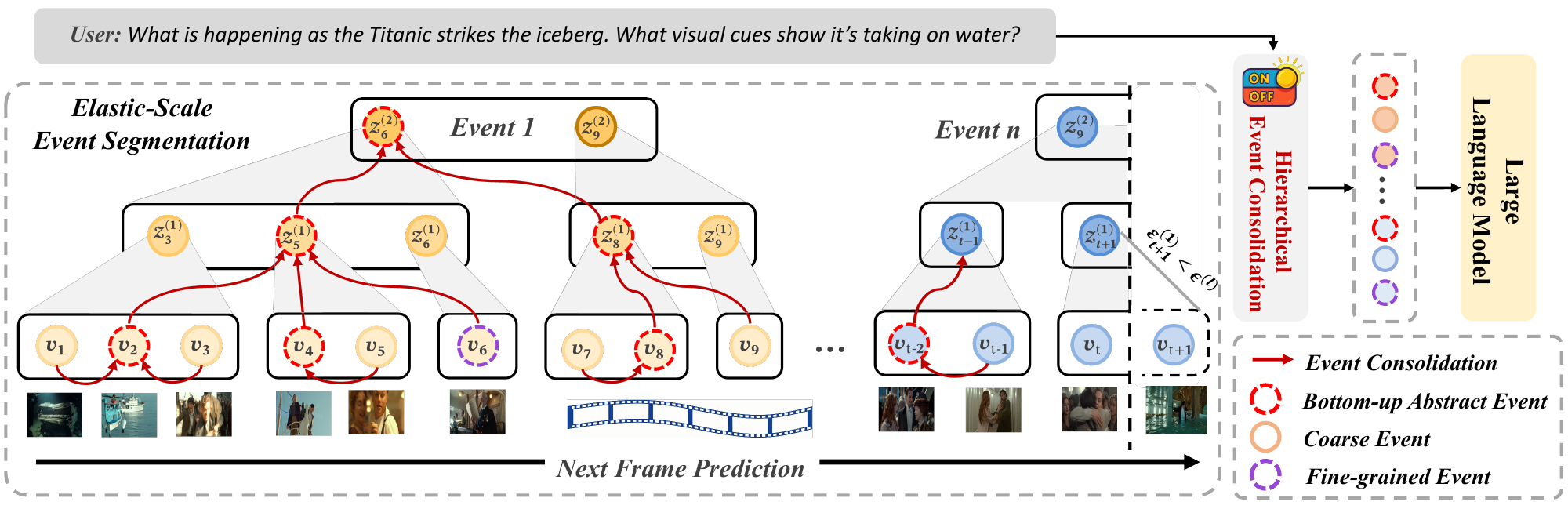}
\vspace{-0.2cm}
\caption{Overview of the proposed VideoScaffold, which comprises two core components: Elastic-Scale Event Segmentation (EES) and Hierarchical Event Consolidation (HEC). As streaming video progresses, a next-frame prediction mechanism continuously refines event boundaries and constructs a hierarchical event structure that adapts to video duration. Upon receiving a user instruction, HEC performs bottom-up summarization along this hierarchy, generating multi-granularity event representations for  subsequent video understanding.}
\label{fig:framework}
\vspace{-0.3cm}
\end{figure*}

To effectively handle this causal, streaming nature, we design VideoScaffold, a visual representation framework that incrementally models temporal dynamics while maintaining long-term coherence. 
As illustrated in Fig.~\ref{fig:framework}, it consists of two complementary components: 
(1) Elastic-scale Event Segmentation (EES), which adaptively refines event boundaries as the video unfolds; and 
(2) Hierarchical Event Consolidation (HEC), which aggregates multi-level representations to capture both detailed and abstract semantics across variable video durations.

\subsection{Elastic-Scale Event Segmentation}

Drawing inspiration from cognitive studies on real-time perception, humans segment continuous sensory input into discrete events for efficient understanding and memory.
Event Segmentation Theory (EST)~\cite{est} explains this phenomenon through predictive mechanisms in the brain, where transient prediction errors between expectation and observation signal event boundaries.
Motivated by this principle, we propose an \textbf{Elastic-Scale Event Segmentation (EES)} framework that incrementally predicts future visual content, compares it with actual observations, and triggers new events when significant deviations occur.
Extending this predictive process across multiple hierarchical levels enables EES to capture elastic event structures that adapt to temporal dynamics over varying video durations.

\noindent\textbf{Base-level abstraction and prediction.}
At the first (bottom) level, the model operates on frame-level visual embeddings.
We define a temporal context window as
\begin{equation}
    \mathcal{Z}^{(1)}_{t} = 
    \{\boldsymbol{v}_{t-m+1}, \ldots, \boldsymbol{v}_{t}\},
    \qquad \boldsymbol{v}_{t} \in \mathbb{R}^{d},
\end{equation}
which denotes a sequence of $m$ consecutive frame embeddings up to time step $t$, assumed to exhibit semantic coherence.  
A learnable abstraction module $\boldsymbol{\Phi}^{(1)}(\cdot; \boldsymbol{\phi}^{(1)})$ maps this temporal window to a compact latent embedding:
\begin{equation}
    \boldsymbol{z}^{(1)}_{t} =
    \boldsymbol{\Phi}^{(1)}(\mathcal{Z}^{(1)}_{t}; \boldsymbol{\phi}^{(1)}),
    \qquad \boldsymbol{z}^{(1)}_{t} \in \mathbb{R}^{d},
\end{equation}
where $\boldsymbol{\phi}^{(1)}$ are the shared parameters of the first-level abstraction layer.
The latent embedding $\boldsymbol{z}^{(1)}_{t}$ serves as a semantic summary of the current event segment and provides the basis for next-frame prediction.  
A prediction module $\boldsymbol{\Psi}^{(1)}(\cdot; \boldsymbol{\psi}^{(1)})$ maps this token to the predicted embedding:
\begin{equation}
    \hat{\boldsymbol{v}}_{t+1} =
    \boldsymbol{\Psi}^{(1)}(\boldsymbol{z}^{(1)}_{t}; \boldsymbol{\psi}^{(1)}),
    \qquad \hat{\boldsymbol{v}}_{t+1} \in \mathbb{R}^{d},
\end{equation}
where $\boldsymbol{\psi}^{(1)}$ denotes parameters of the first prediction layer.

\noindent\textbf{Hierarchical abstraction.}
Higher layers in the hierarchy capture progressively longer temporal dependencies by operating on abstracted tokens from lower layers.  
At level $l>1$, the temporal context is defined as a sequence of $(l\!-\!1)$-level embeddings:
\begin{equation}
    \mathcal{Z}^{(l)}_{t} =
    \{\boldsymbol{z}^{(l-1)}_{t-m_l+1}, \ldots, \boldsymbol{z}^{(l-1)}_{t}\},
\end{equation}
where $m_l$ denotes the temporal window size at level $l$.  
This sequence is abstracted into a higher-level representation:
\begin{equation}
    \boldsymbol{z}^{(l)}_{t} =
    \boldsymbol{\Phi}^{(l)}(\mathcal{Z}^{(l)}_{t}; \boldsymbol{\phi}^{(l)}),
    \qquad \boldsymbol{z}^{(l)}_{t} \in \mathbb{R}^{d}.
\end{equation}

\noindent\textbf{Hierarchical prediction.}
To preserve consistency across levels, the next-step prediction at level $l$ is conditioned on the cumulative representations from all preceding layers:
\begin{equation}
    \hat{\boldsymbol{z}}^{(l)}_{t+1} =
    \boldsymbol{\Psi}^{(l)}(
        \boldsymbol{z}^{(1)}_{t},
        \boldsymbol{z}^{(2)}_{t},
        \ldots,
        \boldsymbol{z}^{(l)}_{t};
        \boldsymbol{\psi}^{(l)}),
    \qquad
    \hat{\boldsymbol{z}}^{(l)}_{t+1} \in \mathbb{R}^{d}.
\end{equation}
Here, $\boldsymbol{\Psi}^{(l)}(\cdot; \boldsymbol{\psi}^{(l)})$ denotes the $l$-th prediction module parameterized by $\boldsymbol{\psi}^{(l)}$.
This hierarchical conditioning enables higher layers to model abstract semantic transitions while remaining grounded in low-level visual cues, facilitating consistent multi-scale temporal reasoning.

\begin{figure*}[!t]
\centering
\includegraphics[width=0.98\textwidth]{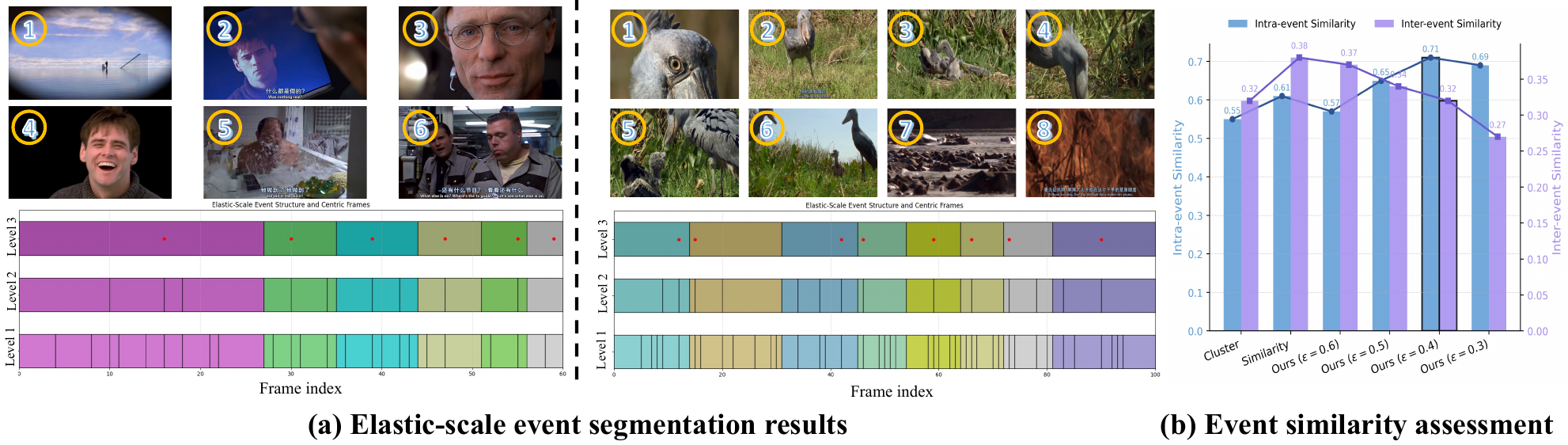}
\vspace{-0.2cm}
\caption{Visualization of the proposed elastic-scale event structure. (a) Layer-wise segmentation results showing how VideoScaffold adaptively refines temporal boundaries and preserves continuity across different video durations (60 and 100 frames).
(b) Comparison of intra-event and inter-event similarity on 100 videos (120 frames) from MLVU, highlighting VideoScaffold’s ability to yield semantically cohesive events and distinct inter-event boundaries compared with clustering- or similarity-based methods.
}
\label{fig:event_segment}
\vspace{-0.3cm}
\end{figure*}

\noindent\textbf{Boundary refinement.}
To enable dynamic hierarchy construction~\cite{streamer}, we quantify the prediction error using cosine distance between the predicted and actual embeddings:
\begin{equation}
    \mathcal{E}^{(l)}_{t+1} =
    1 - \cos\!\left(
        \hat{\boldsymbol{z}}^{(l)}_{t+1},
        \boldsymbol{z}^{(l)}_{t+1}
    \right),
    \qquad \mathcal{E}^{(l)}_{t+1} \in [0,2].
\end{equation}
An event boundary is triggered at level $l$ when the prediction error exceeds a predefined threshold $\epsilon^{(l)}$:
$\mathcal{E}^{(l)}_{t+1} > \epsilon^{(l)}.$
Upon boundary activation, the current latent token $\boldsymbol{z}^{(l)}_{t}$ is finalized and appended to the next-level sequence:
$    \mathcal{Z}^{(l+1)}_{t+1} \leftarrow
    \mathcal{Z}^{(l+1)}_{t} \cup \{\boldsymbol{z}^{(l)}_{t}\}.$
If $\mathcal{E}^{(l)}_{t+1} \le \epsilon^{(l)}$, the current frame continues within the ongoing event, and no new token is emitted.
This adaptive boundary mechanism dynamically refines event granularity according to temporal dynamics, achieving elastic-scale segmentation while preserving the continuity of the video stream (see Fig.~\ref{fig:event_segment}).

\subsection{Hierarchical Event Consolidation}
After performing EES, the video sequence is organized into a hierarchical structure of event segments, forming a tree-like representation across temporal scales.
We denote the highest abstraction level by $L$, and represent an event segment at level $l$ as $\mathcal{S}^{(l)}$.
Each segment consists of a temporally ordered sequence of latent embeddings:
\begin{equation}
    \mathcal{S}^{(l)} =
    \{\boldsymbol{z}^{(l)}_{s}, \boldsymbol{z}^{(l)}_{s+1}, \ldots, \boldsymbol{z}^{(l)}_{t}\},
    \qquad \boldsymbol{z}^{(l)}_{i} \in \mathbb{R}^{d}.
\end{equation}
Upon receiving a user instruction, we perform a bottom-up consolidation process that generates compact and semantically enriched representations for each event.
This process consists of two main components: \emph{essential element identification} and \emph{cross-layer aggregation}, which jointly construct a unified hierarchical representation for downstream video understanding.

\noindent\textbf{Essential element identification.}
Given an event segment $\mathcal{S}^{(l)}$, we first identify the \emph{essential token}, the most semantically informative element within the segment—based on the prediction error computed during segmentation.
Formally, the essential token is defined as:
\begin{equation}
    \boldsymbol{z}_{\mathrm{ess}}^{(l)} = \boldsymbol{z}_{i^{*}}^{(l)},
    \qquad
    i^{*} = \arg\max_{i \in [s,\, t]} \mathcal{E}_{i}^{(l)},
    \label{eq:essential_token}
\end{equation}
where $\mathcal{E}_{i}^{(l)}$ denotes the prediction error at level $l$.
The token functions as a semantic anchor that guides intra-segment contextual aggregation.

\noindent\textbf{Intra-layer aggregation.}
To integrate contextual information from the remaining tokens within the same segment, we employ a cross-attention mechanism in which the essential token serves as the query, while the remaining tokens act as keys and values:
\begin{equation}
    \tilde{\boldsymbol{z}}^{(l)} =
    \operatorname{CrossAttn}\!\left(
        Q = \boldsymbol{z}_{\mathrm{ess}}^{(l)},
        K = V = \mathcal{S}^{(l)} \setminus \{\boldsymbol{z}_{\mathrm{ess}}^{(l)}\}
    \right).
\end{equation}
Applying this operation to all event segments at level $l$ yields a set of summarized event embeddings:
\begin{equation}
    \mathcal{V}^{(l)} =
    \{\tilde{\boldsymbol{z}}_{1}^{(l)}, \tilde{\boldsymbol{z}}_{2}^{(l)}, \ldots, \tilde{\boldsymbol{z}}_{n}^{(l)}\},
    \qquad \tilde{\boldsymbol{z}}_{i}^{(l)} \in \mathbb{R}^{d}.
\end{equation}

\noindent\textbf{Cross-layer aggregation.}
To propagate information upward through the hierarchy, we identify the essential token at level $l{+}1$ using the same criterion defined in Eq.~\ref{eq:essential_token}.
A cross-attention operation is then applied to fuse the summarized embeddings from the lower level:
\begin{equation}
    \tilde{\boldsymbol{z}}^{(l+1)} =
    \operatorname{CrossAttn}\!\left(
        Q = \tilde{\boldsymbol{z}}_{\mathrm{ess}}^{(l+1)},
        K = V = \mathcal{V}^{(l)}
    \right).
\end{equation}
This recursive process is repeated across layers until the top level $L$, producing the final abstract event embedding:
$    \boldsymbol{e}_{\text{abstract}} = \tilde{\boldsymbol{z}}^{(L)}.$

\noindent\textbf{Complementary event embeddings.}
Beyond the abstract representation, two complementary embeddings are derived to capture coarse and fine event semantics.
The \emph{coarse event embedding} summarizes the global context of the top-level segment via average pooling:
\begin{equation}
    \boldsymbol{e}_{\text{coarse}} =
    \frac{1}{|\mathcal{S}^{(L)}|}
    \sum_{\boldsymbol{z} \in \mathcal{S}^{(L)}} \boldsymbol{z}.
\end{equation}
The \emph{fine-grained event embedding} highlights the most discriminative token, identified by the highest prediction error:
\begin{equation}
    \boldsymbol{e}_{\text{fine}} =
    \boldsymbol{z}^{(L)}_{i^{*}},
    \qquad
    i^{*} = \arg\max_{i \in \mathcal{S}^{(L)}} \mathcal{E}^{(L)}_{i}.
\end{equation}
Combined with $\boldsymbol{e}_{\text{abstract}}$, these representations form a unified hierarchical event consolidation that is both semantically expressive and temporally coherent.

\section{Experiments}
\subsection{Implementation Details}
We adopt EVA-CLIP~\cite{eva_clip} as the vision encoder and Vicuna-7B~\cite{vicuna} as the language backbone. 
Training is conducted in two stages: alignment and instruction tuning. 
During both stages, the vision encoder remains frozen, while the language model is frozen in the alignment stage and fine-tuned during instruction tuning. 
The global batch size is set to 128 and 32 for the two stages, with learning rates of $1\times10^{-3}$ and $2\times10^{-5}$, respectively. 
Each stage is trained for one epoch.

\begin{table*}[!t]
\centering
\renewcommand{\arraystretch}{1.0}
\small
\setlength{\tabcolsep}{1.8mm}{
\caption{Comparison of zero-shot performance on video QA benchmarks (MSVD-QA, MSRVTT-QA, ActivityNet-QA) and the long-form video benchmark LV-Bench. VideoScaffold uses 60 frames across all datasets.}
\vspace{-0.1cm}
\begin{tabular}{lc|ccc|ccccccc}
\toprule
\multirow{2}{*}{\textbf{Methods}}  & \multirow{2}{*}{\textbf{Size}} & \textbf{MSVD} & \textbf{MSRVTT} & \textbf{ActivityNet} & \multicolumn{7}{c}{\textbf{LV-Bench}} \\
 & & Acc/Score & Acc/Score & Acc/Score & ER & EU & KIR & TG & Rea & Sum & Overall \\
\midrule
MovieChat~\cite{song2024moviechat} & 7B & 75.2/3.8 & 52.7/2.6 & 45.7/3.4 & 21.3 & 23.1 & 25.9 & 22.3 & 24.0 & 17.2 & 22.5 \\
VideoChatGPT~\cite{video_chatgpt} & 7B & 64.9/3.3 & 49.3/2.8 & 35.2/2.7 & 22.9 & 22.6 & 22.7 & 25.5 & 23.4 & 24.1 & 23.1 \\
Chat-UniVi~\cite{chat_univi} & 7B & 65.0/3.6 & 54.6/3.1 & 45.8/3.2 & - & - & - & - & - & - & - \\
LLaMA-VID~\cite{llama_vid} & 13B & 70.0/3.7 & 58.9/3.3 & 47.5/3.3 & 25.4 & 21.7 & 23.4 & 26.4 & 26.5 & 17.2 & 23.9 \\
LWM~\cite{lwm} & 7B &  55.9/- & 44.1/- & -/- & 24.7 & 24.8 & 26.5 & 28.6 & 30.5 & 22.4 & 25.5 \\
LLaVA-NeXT~\cite{llava_next}* & 7B & -/- & -/- & 53.5/3.2 & 30.1 & 31.2 & 34.1 & 31.4 & 35.0 & 27.6 & 32.2 \\
LLaVA-OV~\cite{llava_onevision}* & 70B & -/- & -/- & 62.3/- & 25.0 & 26.9 & 29.2 & 30.9 & 25.4 & 31.0 & 26.9 \\
\midrule
VideoScaffold (Ours) & 7B & 72.5/3.9 & 58.4/3.3 & 48.9/3.4 & 32.1 & 31.2 & 29.6 & 30.0 & 35.3 & 25.9 & 31.5 \\
LLaVA-SFT (16 frames) & 7B & 68.3/3.4 & 54.7/3.0 & 44.5/3.2 & 23.1 & 25.4 & 23.7 & 20.4 & 22.8 & 21.2 & 23.1 \\
LLaVA-SFT (16 frames) + Ours & 7B & 69.5/3.6 & 55.7/3.2 & 45.9/3.2 & 24.7 & 26.6 & 25.3 & 26.1 & 29.7 & 27.7 & 24.5 \\
LLaVA-SFT + Ours (60 frames) & 7B & 70.8/3.6 & 56.3/3.2 & 46.9/3.3 & 29.2 & 28.1 & 28.5 & 28.2 & 33.3 & 29.3 & 29.1 \\
\bottomrule
\label{tab:lvbench}
\vspace{-0.5cm}
\end{tabular}}
\end{table*}

\begin{table*}[!t]
\centering
\small
\renewcommand{\arraystretch}{1.0}
\setlength{\tabcolsep}{2.5mm}{
\caption{Performance comparison on long video benchmark MLVU, including the holistic, single-detail, and multi-detail tasks. 
}
\vspace{-0.1cm}
\label{tab:mlvu}
\begin{tabular}{l|c|cccc|ccc|cc|c|c}
\toprule
\multirow{2}{*}{\textbf{Methods}} & \multirow{2}{*}{\textbf{Frames}} & \multicolumn{4}{c|}{\textbf{Holistic}} & \multicolumn{3}{c|}{\textbf{Single Detail}} & \multicolumn{2}{c|}{\textbf{Multi Detail}} & \multirow{2}{*}{\textbf{M-Avg}} & \multirow{2}{*}{\textbf{G-Avg}} \\
& & TR & AR & VS & NQA & ER & PQA & SSC & AO & AC & & \\
\midrule
Video-ChatGPT~\cite{video_chatgpt} & 100  & 26.9 & 24.0 & 2.31 & 40.3 & 42.0 & 29.9 & 5.48 & 25.1 & 31.1 & 31.3 & 3.90 \\
VideoChat2~\cite{li2024mvbench} & 16  & 74.6 & 51.5 & 2.57 & 42.0 & 47.4 & 43.8 & 5.04 & 22.8 & 29.6 & 44.5 & 3.81 \\
Video-LLaVA~\cite{video_llava} & 8  & 71.6 & 57.0 & 2.43 & 53.2 & 45.2 & 48.4 & 5.25 & 20.1 & 35.9 & 47.3 & 3.84 \\
MovieChat~\cite{song2024moviechat} & 2048 & 29.5 & 25.0 & 2.33 & 24.2 & 24.7 & 25.8 & 3.23 & 28.6 & 22.8 & 25.8 & 2.78 \\
LLaMA-VID~\cite{llama_vid} & 1 & 50.8 & 34.5 & 3.22 & 30.1 & 32.7 & 32.5 & 5.22 & 23.9 & 27.8 & 33.2 & 4.22 \\
MA-LMM~\cite{ma_llm} & 1000 & 51.9 & 35.5 & 2.12 & 43.1 & 38.9 & 35.8 & 4.80 & 25.1 & 24.3 & 36.4 & 3.46 \\
LLaVA-OV~\cite{llava_onevision}* & 32 & 83.5 & 56.4 & 3.75 & 46.7 & 58.4 & 58.0 & 5.09 & 35.7 & 23.3 & 51.7 & 4.42 \\
\midrule
VideoScaffold (Ours) & 60 & 73.9 & 55.0 & 3.58 & 55.2 & 48.6 & 52.1 & 5.16 & 24.3 & 37.4 & 49.5 & 4.37 \\
LLaVA-SFT & 16 & 71.8 & 51.0 & 2.84 & 41.9 & 44.6 & 42.7 & 4.77 & 28.2 & 28.1 & 44.0 & 3.81 \\
LLaVA-SFT + Ours & 60 & 75.0 & 56.5 & 3.05 & 43.9 & 46.4 & 47.7 & 4.92 & 28.5 & 24.8 & 46.1 & 3.99 \\
\bottomrule
\end{tabular}}
\vspace{-0.4cm}
\end{table*}

We follow a joint image--video training paradigm~\cite{llava,chat_univi} to enhance multimodal understanding. 
For alignment, we employ the LLaVA-filtered CC3M~\cite{sharma2018conceptual} and WebVid-2.5M~\cite{bain2021frozen} datasets for image and video captioning. 
For instruction tuning, we use LLaVA-665K~\cite{llava} and the ActivityNet-derived dataset from~\cite{video_chatgpt} for image- and video-based question answering.

\subsection{Benchmarks}
\textbf{Open-Ended QA.} 
We evaluate zero-shot performance on three open-ended video QA benchmarks: 
ActivityNet-QA~\cite{activitynet}, MSVD-QA~\cite{msvd}, and MSRVTT-QA~\cite{msrvtt}, which assess fine-grained temporal and visual understanding.

\begin{table*}[!t]
\centering
\renewcommand{\arraystretch}{1.0}
\setlength{\tabcolsep}{2.5mm}{
\caption{Performance comparison on long video benchmark, VideoMME,  across short, medium, and long video durations, evaluated under both “without subtitles” and “with subtitles” settings. * denotes models trained on substantially larger datasets than ours.}
\label{tab:videomme}
\begin{tabular}{lcc|cccc|cccc}
\toprule
\multirow{2}{*}{\textbf{Methods}} & \multirow{2}{*}{\textbf{LLM Size}} & \multirow{2}{*}{\textbf{Frames}} & \multicolumn{4}{c|}{\textbf{VideoMME w/o sub.}} & \multicolumn{4}{c}{\textbf{VideoMME w/ sub.}} \\
 & & & \textbf{Short} & \textbf{Mid} & \textbf{Long} & \textbf{All} & \textbf{Short} & \textbf{Mid} & \textbf{Long} & \textbf{All} \\
\midrule
LLaMA-VID~\cite{llama_vid} & 7B & 1fps & - & - & - & 25.9 & - & - & - & - \\
Video-LLaVA~\cite{video_llava} & 7B & 8 & 45.3 & 38.0 & 36.2 & 39.9 & 46.1 & 40.7 & 38.1 & 41.6 \\
VideoChat2~\cite{li2024mvbench} & 7B & 16 & 48.3 & 36.3 & 35.0 & 39.5 & 52.8 & 39.4 & 39.2 & 43.8 \\
Chat-UniVi\cite{chat_univi} & 7B & 64 & 45.7 & 40.3 & 35.8 & 40.6 & 51.2 & 44.6 & 41.8 & 45.9 \\
VideoLLaMA2~\cite{Videollama2}* & 7B & 16 & 56.0 & 45.4 & 42.1 & 47.9 & - & - & - & - \\
LLaVA-OneVision~\cite{llava_onevision}* & 7B & 32 & 70.1 & {56.4} & {48.9} & {58.5} & {75.8} & {58.4} & {51.6} & {61.9} \\
\midrule
VideoScaffold (Ours) & 7B & 60 & 47.2 & 43.6 & 39.2 & 43.3 & 52.9 & 45.5 & 43.2 & 47.2 \\
LLaVA-SFT & 7B & 16 & 40.2 & 34.0 & 32.3 & 35.5 & 42.6 & 42.3 & 37.9 & 40.9  \\
LLaVA-SFT + Ours & 7B & 60 & 42.1 & 38.4 & 37.7 & 39.4 & 45.3 & 44.2 & 39.5 & 43.0 \\
\bottomrule
\end{tabular}}
\end{table*}

\begin{table*}[!t]
\centering
\small
\renewcommand{\arraystretch}{1.0}
\setlength{\tabcolsep}{2.2mm}{
\caption{Performance comparison on StreamingBench on Real-Time Visual Understanding. 
% * denotes models trained on substantially larger datasets than ours.
}
\label{tab:streamingBench}
\small
\begin{tabular}{l|c|c|ccccccccccc}
\toprule
\textbf{Models} & \textbf{Size} & \textbf{Frames} & \textbf{OP} & \textbf{CR} & \textbf{CS} & \textbf{ATP} & \textbf{EU} & \textbf{TR} & \textbf{PR} & \textbf{SU} & \textbf{ACP} & \textbf{CT} & \textbf{All} \\
\midrule
\multicolumn{14}{c}{\textit{Human}} \\
\midrule
Human$^\ddagger$ & - & - & 89.5 & 92.0 & 93.6 & 91.5 & 95.7 & 92.5 & 88.0 & 88.8 & 89.7 & 91.3 & 91.5 \\
\midrule
\multicolumn{14}{c}{\textit{Open-Source Video MLLMs}} \\
\midrule
LLaVA-OneVision~\cite{llava_onevision}* & 7B & 32 & 80.4 & 74.2 & 76.0 & 80.7 & 72.7 & 71.6 & 67.6 & 65.5 & 65.7 & 45.1 & 71.1 \\
Qwen2-VL~\cite{qwen2_vl}* & 7B & 0.2-1 fps & 75.2 & 82.8 & 73.2 & 77.5 & 68.3 & 71.0 & 72.2 & 61.2 & 61.5 & 46.1 & 69.0 \\
MiniCPM-V2.6~\cite{Minicpm}* & 8B & 32 & 71.9 & 71.1 & 77.9 & 75.8 & 64.6 & 65.7 & 70.4 & 56.1 & 62.3 & 53.4 & 67.4 \\
LLaVA-NeXT~\cite{llava_next}* & 32B & 64 & 78.2 & 70.3 & 73.8 & 76.8 & 63.4 & 69.8 & 57.4 & 56.1 & 64.3 & 38.9 & 67.0 \\
InternVL2\cite{internvl2}* & 8B & 16 & 68.1 & 60.9 & 69.4 & 77.1 & 67.7 & 62.9 & 59.3 & 53.3 & 55.0 & 56.5 & 63.7 \\
Kangaroo~\cite{liu2024kangaroo}* & 7B & 64 & 71.1 & 84.4 & 70.7 & 73.2 & 60.0 & 61.7 & 45.9 & 55.7 & 62.0 & 38.6 & 62.1 \\
LongVA~\cite{longva}* & 7B & 128 & 70.0 & 63.3 & 61.2 & 70.9 & 62.7 & 59.5 & 61.1 & 53.7 & 54.7 & 34.7 & 60.0 \\
VILA-1.5~\cite{vila2}* & 8B & 14 & 53.7 & 49.2 & 71.0 & 56.9 & 53.4 & 53.9 & 54.6 & 48.8 & 50.1 & 17.6 & 52.3 \\
Video-LLaMA2~\cite{Videollama2}* & 7B & 32 & 55.9 & 55.5 & 57.4 & 58.2 & 52.8 & 43.6 & 39.8 & 42.7 & 45.6 & 35.2 & 49.5 \\
\midrule
\multicolumn{14}{c}{\textit{Streaming MLLMs}} \\
\midrule
Flash-VStream~\cite{flash_vstream} & 7B & - & 25.9 & 43.6 & 24.9 & 23.9 & 27.3 & 13.1 & 18.5 & 25.2 & 23.9 & 48.7 & 23.2 \\
VideoLLM-online~\cite{chen2024videollm}* & 8B & 2 fps & 39.1 & 40.1 & 34.5 & 31.1 & 46.0 & 32.4 & 31.5 & 34.2 & 42.5 & 27.9 & 36.0 \\
  & 7B & 0.5 fps & 41.9 & 37.6 & 42.6 & 40.5 & 51.5 & 35.5 & 37.7 & 44.9 & 42.7 & 34.6 & 41.0 \\
LLaVA-SFT + Ours & 7B & 0.5 fps & 32.6 & 27.1 & 33.0 & 30.8 & 41.2 & 25.0 & 27.5 & 34.6 & 33.1 & 24.3 & 31.1 \\
\bottomrule
\end{tabular}}
\vspace{-0.2cm}
\end{table*}

\textbf{Long-Video Understanding.} 
To evaluate long-form comprehension, we adopt three multiple-choice benchmarks: LVBench~\cite{Lvbench}, VideoMME~\cite{video_mme}, and MLVU~\cite{mlvu}. 
LVBench includes 1.5K QA pairs from videos exceeding 60 minutes, 
VideoMME consists of 900 videos with 2.7K QA pairs spanning diverse durations, and MLVU features both multiple-choice and open-ended tasks on videos ranging from 3 minutes to 2 hours.

\textbf{Streaming Evaluation.} 
We further adopt StreamingBench~\cite{streamingbench}, 
a recent benchmark specifically designed for evaluating streaming video understanding in MLLMs. 
We utilize its real-time visual comprehension subset to assess the streaming adaptability of our model.

\subsection{Main Results}
\textbf{Short-Video Understanding.}
As shown in Table~\ref{tab:lvbench}, VideoScaffold consistently outperforms existing baselines across three benchmarks under comparable data scales. 
Despite employing a smaller 7B backbone, it surpasses LLaMA-VID (13B), demonstrating superior efficiency and scalability. 
Integrating VideoScaffold with LLaVA further improves performance, achieving gains of 2.5\%, 1.6\%, and 1.4\% in accuracy on the respective benchmarks. 
These results highlight the effectiveness of our elastic-scale event representation, which captures temporally coherent and semantically expressive structures beyond static sampling.

\textbf{Long-Video Understanding.}
To further evaluate the effectiveness of our elastic-scale visual hierarchies, we conduct comprehensive comparisons across three long-form video understanding benchmarks: LV-Bench, MLVU, and VideoMME, as reported in Tables~\ref{tab:lvbench}, \ref{tab:mlvu}, and \ref{tab:videomme}, respectively. 
For LV-Bench, which comprises videos exceeding one hour in duration, our model outperforms all 7B-scale baselines and performs comparably to LLaVA-OneVision, despite the latter employing a significantly larger 70B language model. 
In the MLVU benchmark, although several competing approaches such as MovieChat (2048 frames) and MA-LMM (1000 frames) utilize substantially more visual tokens, VideoScaffold achieves superior results using only 60 frames, surpassing all baselines except LLaVA-OneVision. 
On VideoMME, our model attains the highest performance among methods trained with comparable data scales. 
These results collectively demonstrate the strong scalability and robustness of our approach for long-duration video comprehension tasks.

Serving as a plug-and-play module, VideoScaffold endows LLaVA with enhanced capability for long-video understanding. 
It consistently delivers stable performance improvements across all benchmarks, achieving gains of 6.0, 2.1, and 3.9, respectively, while maintaining generalization and scalability across diverse benchmarks.

\textbf{Streaming Video Understanding.}
As a recent benchmark for real-time video comprehension, StreamingBench~\cite{streamingbench} provides a comprehensive evaluation of MLLMs in continuous streaming scenarios. 
We report the results on its real-time visual understanding subset in Table~\ref{tab:streamingBench}. 
The results show that our VideoScaffold achieves the highest average performance among existing streaming MLLMs, outperforming VideoLLM-online even though the latter is trained on a larger-scale dataset. 
This demonstrates that our prediction-guided event segmentation effectively captures temporal continuity while maintaining computational efficiency under causal constraints, leading to more stable understanding in streaming environments.
Furthermore, integrating VideoScaffold with LLaVA substantially extends its capability from static image or offline video understanding to dynamic, real-time streaming contexts, exhibiting strong adaptability, and generalization across continuously evolving multimodal inputs.

\subsection{Ablation Studies}
\label{sec:abaltion}
\textbf{Elastic-Scale Event Segmentation.} 
We conduct an ablation study to evaluate the effectiveness of the elastic hierarchy in the proposed EES module. 
To this end, we restrict the segmentation process to a single fine-grained level, disabling hierarchical evolution. 
As shown in Table~\ref{tab:ablation_seg}, performance on short videos (a few seconds) remains largely unaffected, as hierarchical abstraction is unnecessary in such cases. 
In contrast, long videos (several minutes) exhibit notable degradation due to fragmented segmentation and increased redundancy, highlighting the importance of dynamically adapting event granularity to video duration. 
We further examine the execution space of the next-frame prediction mechanism by comparing operations in pixel and latent spaces, as well as across class and patch tokens. 
Results show that prediction in the latent space using patch tokens achieves the best performance, confirming the advantage of compact semantic modeling over raw pixel-level prediction.

\begin{table}[!t]
\centering
\small
\renewcommand{\arraystretch}{1.0}
\setlength{\tabcolsep}{1.0mm}{
\captionof{table}{Ablation about the elastic-scale event segmentation.}
\begin{tabular}{ccc|cccc}
\toprule
\textbf{Elastic} & \textbf{Space} & \textbf{Token} & \textbf{ANet} & \textbf{MSVD} & \textbf{MSRVTT} & \textbf{LV-Bench} \\
$\times$ & \emph{Latent} & \emph{Patch} & 44.3 & 71.0 & 56.8 & 27.3 \\
$\checkmark$ & \emph{Pixel} & \emph{-} & 45.5 & 70.4 & 56.1 & 28.2 \\
$\checkmark$ & \emph{Latent} & \emph{Cls} & 45.2 & 70.2 & 55.5 & 27.8 \\
$\checkmark$ & \textbf{\emph{Latent}} & \textbf{\emph{Patch}} & \textbf{48.9} & \textbf{72.5} & \textbf{58.4} & \textbf{31.5} \\
\bottomrule
\label{tab:ablation_seg}
\end{tabular}}
\vspace{-0.6cm}
\end{table}

\textbf{Event Layers and Threshold.}
We examine two key factors in the EES module: the number of event layers and the boundary threshold $\varepsilon$. 
As shown in Table ~\ref{tab:ablation_layer}, increasing the layer count from one to three steadily improves performance by capturing longer temporal dependencies and higher-level semantics. 
However, further deepening brings marginal gains while increasing inference latency, suggesting that excessive abstraction introduces redundancy. 
Meanwhile, the segmentation threshold $\varepsilon$ controls event granularity: smaller values cause over-fragmented events, disrupting temporal coherence, whereas larger values merge distinct scenes, reducing discriminability. 
Performance peaks around $\varepsilon=0.4$, achieving a balanced trade-off between temporal compactness and semantic separation. 
These results confirm that a three-layer hierarchy with a moderate threshold provides the best balance between representation richness and efficiency.

\begin{figure}[!t]
\centering
\includegraphics[width=0.46\textwidth]{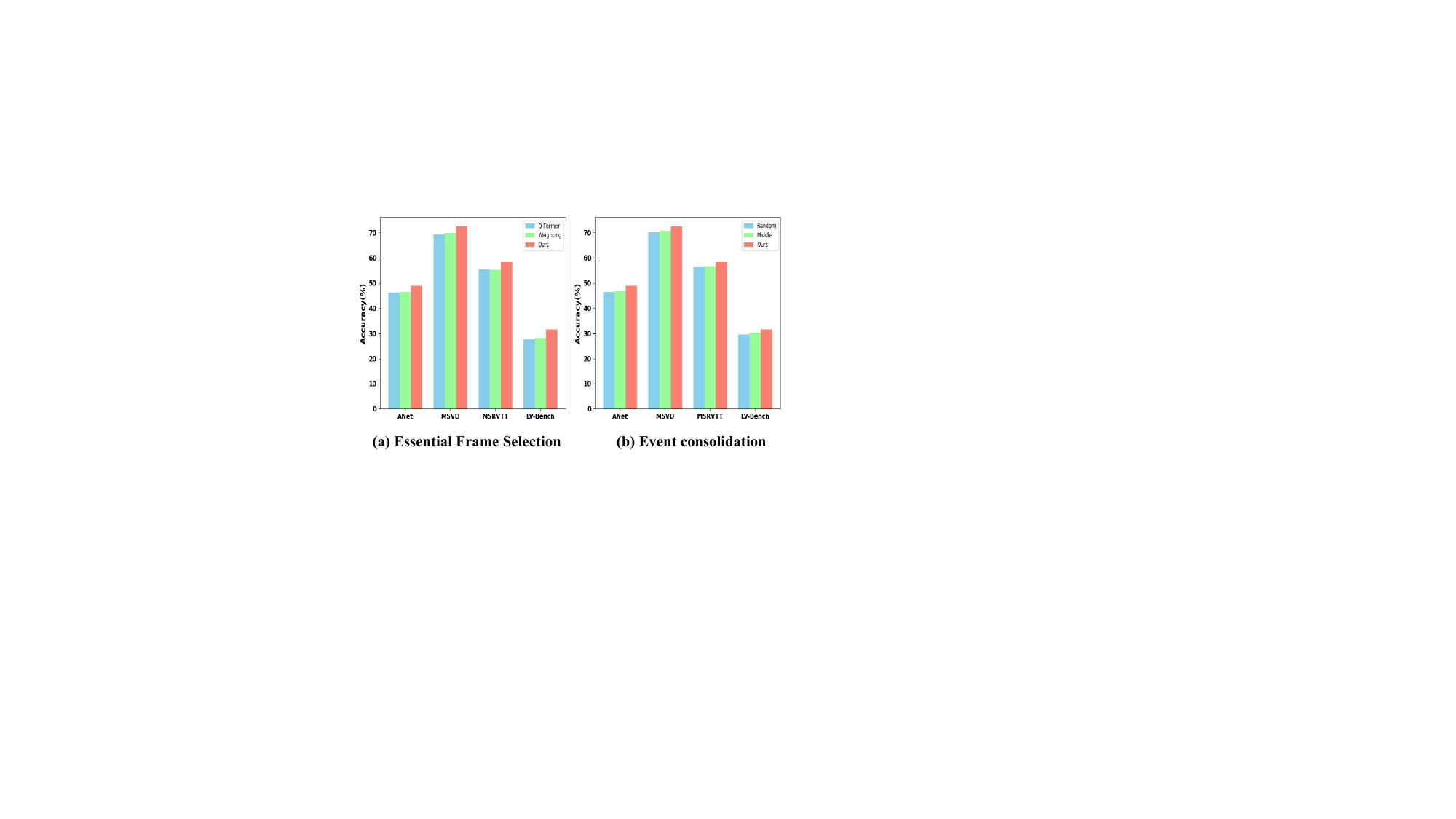}
\vspace{-0.2cm}
\caption{Ablation on essential frame selection and consolidation.}
\label{fig:ablation_centric}
\vspace{-0.2cm}
\end{figure}

\textbf{Hierarchical Event Consolidation.} 
As shown in Fig.~\ref{fig:ablation_centric}(a), we evaluate aggregation strategies within the proposed HEC module, comparing three alternatives: Q-former, distance-based weighting, and our cross-attention approach. 
The Q-former employs a shared learnable query token across all events, yet its lack of event-specific adaptability leads to suboptimal performance. 
The distance-based weighting method aggregates features solely by temporal proximity, neglecting spatial redundancy and producing coarse event representations. 
In contrast, our cross-attention mechanism leverages spatial tokens from the essential frame to selectively attend to event-relevant regions, enabling more accurate and context-aware summarization.

\textbf{Essential Element Selection.} 
Built upon the elastic event structure, the proposed HEC module selects one representative frame per sub-event and performs bottom-up aggregation. 
We evaluate four strategies: random sampling, middle frame, and frames with maximum prediction error (default in this paper). 
As shown in Fig.~\ref{fig:ablation_centric}(b), random and middle-frame selections yield inferior results, misaligning with informative content and leading to suboptimal aggregation. 
In contrast, selecting the frame with the highest prediction error achieves the best performance by effectively capturing visual changes as informative anchors.

\begin{table}[!t]
\centering
\small
\renewcommand{\arraystretch}{1.0}
\setlength{\tabcolsep}{1.0mm}{
\captionof{table}{Ablation about the event layer and threshold.}
\vspace{-0.2cm}
\begin{tabular}{cc|cccc}
\toprule
\textbf{Layers} & \textbf{threshold} & \textbf{ANet} & \textbf{MSVD} & \textbf{MSRVTT} & \textbf{LV-Bench} \\
$3$ & 0.3 & 48.1 & 71.3 & 57.6 & 31.2 \\
$\textbf{3}$ & \textbf{0.4} & \textbf{48.9} & \textbf{72.5} & \textbf{58.4} & \textbf{31.5} \\
$3$ & 0.5 & 47.6 & 70.8 & 56.6 & 30.8 \\
$3$ & 0.6 & 47.0 & 69.7 & 56.2 & 30.3 \\
$2$ & 0.4 & 44.3 & 68.6 & 53.9 & 27.7 \\
$4$ & 0.4 & 48.4 & 72.9 & 58.7 & 31.6 \\
\bottomrule
\label{tab:ablation_layer}
\end{tabular}}
\vspace{-0.6cm}
\end{table}

\section{Conclusion}
In this paper, we present VideoScaffold, a dynamic video representation framework for streaming video understanding with multimodal large language models (MLLMs). 
Unlike offline approaches that struggle with continuous and unbounded inputs, VideoScaffold introduces elastic-scale event segmentation and hierarchical event consolidation to adaptively organize video streams while preserving essential semantics. 
Through a next-frame prediction mechanism and hierarchical bottom-up consolidation, it effectively captures both fine-grained visual cues and high-level abstractions. 
Experimental results show that VideoScaffold consistently outperforms existing methods across both offline and streaming benchmarks, achieving efficient video comprehension. 
Its modular design further allows seamless integration with existing MLLMs, paving the way toward scalable and real-time multimodal video understanding.
{
    \small
    \bibliographystyle{ieeenat_fullname}
    \bibliography{main}

\begin{thebibliography}{61}
\providecommand{\natexlab}[1]{#1}
\providecommand{\url}[1]{\texttt{#1}}
\expandafter\ifx\csname urlstyle\endcsname\relax
  \providecommand{\doi}[1]{doi: #1}\else
  \providecommand{\doi}{doi: \begingroup \urlstyle{rm}\Url}\fi

\bibitem[Achiam et~al.(2023)Achiam, Adler, Agarwal, Ahmad, Akkaya, Aleman, Almeida, Altenschmidt, Altman, Anadkat, et~al.]{achiam2023gpt}
Josh Achiam, Steven Adler, Sandhini Agarwal, Lama Ahmad, Ilge Akkaya, Florencia~Leoni Aleman, Diogo Almeida, Janko Altenschmidt, Sam Altman, Shyamal Anadkat, et~al.
\newblock Gpt-4 technical report.
\newblock \emph{arXiv preprint arXiv:2303.08774}, 2023.

\bibitem[Alayrac et~al.(2022)Alayrac, Donahue, Luc, Miech, Barr, Hasson, Lenc, Mensch, Millican, Reynolds, et~al.]{flamingo}
Jean-Baptiste Alayrac, Jeff Donahue, Pauline Luc, Antoine Miech, Iain Barr, Yana Hasson, Karel Lenc, Arthur Mensch, Katherine Millican, Malcolm Reynolds, et~al.
\newblock Flamingo: a visual language model for few-shot learning.
\newblock \emph{Advances in neural information processing systems}, 35:\penalty0 23716--23736, 2022.

\bibitem[Bain et~al.(2021)Bain, Nagrani, Varol, and Zisserman]{bain2021frozen}
Max Bain, Arsha Nagrani, G{\"u}l Varol, and Andrew Zisserman.
\newblock Frozen in time: A joint video and image encoder for end-to-end retrieval.
\newblock In \emph{Proceedings of the IEEE/CVF international conference on computer vision}, pages 1728--1738, 2021.

\bibitem[Caba~Heilbron et~al.(2015)Caba~Heilbron, Escorcia, Ghanem, and Carlos~Niebles]{activitynet}
Fabian Caba~Heilbron, Victor Escorcia, Bernard Ghanem, and Juan Carlos~Niebles.
\newblock Activitynet: A large-scale video benchmark for human activity understanding.
\newblock In \emph{Proceedings of the ieee conference on computer vision and pattern recognition}, pages 961--970, 2015.

\bibitem[Cai et~al.(2024)Cai, Cao, Chen, Chen, Chen, Chen, Chen, Chen, Chen, Chu, et~al.]{cai2024internlm2}
Zheng Cai, Maosong Cao, Haojiong Chen, Kai Chen, Keyu Chen, Xin Chen, Xun Chen, Zehui Chen, Zhi Chen, Pei Chu, et~al.
\newblock Internlm2 technical report.
\newblock \emph{arXiv preprint arXiv:2403.17297}, 2024.

\bibitem[Chen and Dolan(2011)]{msvd}
David Chen and William~B Dolan.
\newblock Collecting highly parallel data for paraphrase evaluation.
\newblock In \emph{Proceedings of the 49th annual meeting of the association for computational linguistics: human language technologies}, pages 190--200, 2011.

\bibitem[Chen et~al.(2024{\natexlab{a}})Chen, Lv, Wu, Lin, Song, Gao, Liu, Gao, Mao, and Shou]{chen2024videollm}
Joya Chen, Zhaoyang Lv, Shiwei Wu, Kevin~Qinghong Lin, Chenan Song, Difei Gao, Jia-Wei Liu, Ziteng Gao, Dongxing Mao, and Mike~Zheng Shou.
\newblock Videollm-online: Online video large language model for streaming video.
\newblock In \emph{Proceedings of the IEEE/CVF Conference on Computer Vision and Pattern Recognition}, pages 18407--18418, 2024{\natexlab{a}}.

\bibitem[Chen et~al.(2024{\natexlab{b}})Chen, Wu, Wang, Su, Chen, Xing, Zhong, Zhang, Zhu, Lu, et~al.]{internvl}
Zhe Chen, Jiannan Wu, Wenhai Wang, Weijie Su, Guo Chen, Sen Xing, Muyan Zhong, Qinglong Zhang, Xizhou Zhu, Lewei Lu, et~al.
\newblock Internvl: Scaling up vision foundation models and aligning for generic visual-linguistic tasks.
\newblock In \emph{Proceedings of the IEEE/CVF conference on computer vision and pattern recognition}, pages 24185--24198, 2024{\natexlab{b}}.

\bibitem[Chen et~al.(2024{\natexlab{c}})Chen, Wu, Wang, Su, Chen, Xing, Zhong, Zhang, Zhu, Lu, et~al.]{internvl2}
Zhe Chen, Jiannan Wu, Wenhai Wang, Weijie Su, Guo Chen, Sen Xing, Muyan Zhong, Qinglong Zhang, Xizhou Zhu, Lewei Lu, et~al.
\newblock Internvl: Scaling up vision foundation models and aligning for generic visual-linguistic tasks.
\newblock In \emph{Proceedings of the IEEE/CVF conference on computer vision and pattern recognition}, pages 24185--24198, 2024{\natexlab{c}}.

\bibitem[Cheng et~al.(2024)Cheng, Leng, Zhang, Xin, Li, Chen, Zhu, Zhang, Luo, Zhao, et~al.]{Videollama2}
Zesen Cheng, Sicong Leng, Hang Zhang, Yifei Xin, Xin Li, Guanzheng Chen, Yongxin Zhu, Wenqi Zhang, Ziyang Luo, Deli Zhao, et~al.
\newblock Videollama 2: Advancing spatial-temporal modeling and audio understanding in video-llms.
\newblock \emph{arXiv preprint arXiv:2406.07476}, 2024.

\bibitem[Chiang et~al.(2023)Chiang, Li, Lin, Sheng, Wu, Zhang, Zheng, Zhuang, Zhuang, Gonzalez, et~al.]{vicuna}
Wei-Lin Chiang, Zhuohan Li, Zi Lin, Ying Sheng, Zhanghao Wu, Hao Zhang, Lianmin Zheng, Siyuan Zhuang, Yonghao Zhuang, Joseph~E Gonzalez, et~al.
\newblock Vicuna: An open-source chatbot impressing gpt-4 with 90\%* chatgpt quality, march 2023.
\newblock \emph{URL https://lmsys. org/blog/2023-03-30-vicuna}, 3\penalty0 (5), 2023.

\bibitem[Devlin et~al.(2019)Devlin, Chang, Lee, and Toutanova]{devlin2019bert}
Jacob Devlin, Ming-Wei Chang, Kenton Lee, and Kristina Toutanova.
\newblock Bert: Pre-training of deep bidirectional transformers for language understanding.
\newblock In \emph{Proceedings of the 2019 conference of the North American chapter of the association for computational linguistics: human language technologies, volume 1 (long and short papers)}, pages 4171--4186, 2019.

\bibitem[Fang et~al.(2023)Fang, Wang, Xie, Sun, Wu, Wang, Huang, Wang, and Cao]{eva_clip}
Yuxin Fang, Wen Wang, Binhui Xie, Quan Sun, Ledell Wu, Xinggang Wang, Tiejun Huang, Xinlong Wang, and Yue Cao.
\newblock Eva: Exploring the limits of masked visual representation learning at scale.
\newblock In \emph{Proceedings of the IEEE/CVF conference on computer vision and pattern recognition}, pages 19358--19369, 2023.

\bibitem[Fang et~al.(2024)Fang, Zhu, Lu, Wang, Molchanov, Kautz, Cho, Pavone, Han, and Yin]{vila2}
Yunhao Fang, Ligeng Zhu, Yao Lu, Yan Wang, Pavlo Molchanov, Jan Kautz, Jang~Hyun Cho, Marco Pavone, Song Han, and Hongxu Yin.
\newblock Vila2: Vila augmented vila.
\newblock \emph{arXiv preprint arXiv:2407.17453}, 2024.

\bibitem[Fu et~al.(2024)Fu, Dai, Luo, Li, Ren, Zhang, Wang, Zhou, Shen, Zhang, et~al.]{video_mme}
Chaoyou Fu, Yuhan Dai, Yongdong Luo, Lei Li, Shuhuai Ren, Renrui Zhang, Zihan Wang, Chenyu Zhou, Yunhang Shen, Mengdan Zhang, et~al.
\newblock Video-mme: The first-ever comprehensive evaluation benchmark of multi-modal llms in video analysis.
\newblock \emph{arXiv preprint arXiv:2405.21075}, 2024.

\bibitem[Grattafiori et~al.(2024)Grattafiori, Dubey, Jauhri, Pandey, Kadian, Al-Dahle, Letman, Mathur, Schelten, Vaughan, et~al.]{grattafiori2024llama}
Aaron Grattafiori, Abhimanyu Dubey, Abhinav Jauhri, Abhinav Pandey, Abhishek Kadian, Ahmad Al-Dahle, Aiesha Letman, Akhil Mathur, Alan Schelten, Alex Vaughan, et~al.
\newblock The llama 3 herd of models.
\newblock \emph{arXiv preprint arXiv:2407.21783}, 2024.

\bibitem[Han et~al.(2024{\natexlab{a}})Han, Huang, Shi, Zhuo, Su, Zhang, Zhou, Qi, Liao, and Liu]{videoespresso}
Songhao Han, Wei Huang, Hairong Shi, Le Zhuo, Xiu Su, Shifeng Zhang, Xu Zhou, Xiaojuan Qi, Yue Liao, and Si Liu.
\newblock Videoespresso: A large-scale chain-of-thought dataset for fine-grained video reasoning via core frame selection.
\newblock \emph{arXiv preprint arXiv:2411.14794}, 2024{\natexlab{a}}.

\bibitem[Han et~al.(2024{\natexlab{b}})Han, Guo, Pan, Liu, Guan, and Yang]{dynfocus}
Yudong Han, Qingpei Guo, Liyuan Pan, Liu Liu, Yu Guan, and Ming Yang.
\newblock Dynfocus: Dynamic cooperative network empowers llms with video understanding.
\newblock \emph{arXiv preprint arXiv:2411.12355}, 2024{\natexlab{b}}.

\bibitem[He et~al.(2024)He, Li, Jang, Jia, Cao, Shah, Shrivastava, and Lim]{ma_llm}
Bo He, Hengduo Li, Young~Kyun Jang, Menglin Jia, Xuefei Cao, Ashish Shah, Abhinav Shrivastava, and Ser-Nam Lim.
\newblock Ma-lmm: Memory-augmented large multimodal model for long-term video understanding.
\newblock In \emph{Proceedings of the IEEE/CVF Conference on Computer Vision and Pattern Recognition}, pages 13504--13514, 2024.

\bibitem[Jin et~al.(2024)Jin, Takanobu, Zhang, Cao, and Yuan]{chat_univi}
Peng Jin, Ryuichi Takanobu, Wancai Zhang, Xiaochun Cao, and Li Yuan.
\newblock Chat-univi: Unified visual representation empowers large language models with image and video understanding.
\newblock In \emph{Proceedings of the IEEE/CVF Conference on Computer Vision and Pattern Recognition}, pages 13700--13710, 2024.

\bibitem[Li et~al.(2024{\natexlab{a}})Li, Zhang, Guo, Zhang, Li, Zhang, Zhang, Zhang, Li, Liu, et~al.]{llava_onevision}
Bo Li, Yuanhan Zhang, Dong Guo, Renrui Zhang, Feng Li, Hao Zhang, Kaichen Zhang, Peiyuan Zhang, Yanwei Li, Ziwei Liu, et~al.
\newblock Llava-onevision: Easy visual task transfer.
\newblock \emph{arXiv preprint arXiv:2408.03326}, 2024{\natexlab{a}}.

\bibitem[Li et~al.(2024{\natexlab{b}})Li, Zhang, Zhang, Zhang, Li, Li, Ma, and Li]{llava_interleave}
Feng Li, Renrui Zhang, Hao Zhang, Yuanhan Zhang, Bo Li, Wei Li, Zejun Ma, and Chunyuan Li.
\newblock Llava-next-interleave: Tackling multi-image, video, and 3d in large multimodal models.
\newblock \emph{arXiv preprint arXiv:2407.07895}, 2024{\natexlab{b}}.

\bibitem[Li et~al.(2023{\natexlab{a}})Li, Li, Savarese, and Hoi]{blip2}
Junnan Li, Dongxu Li, Silvio Savarese, and Steven Hoi.
\newblock Blip-2: Bootstrapping language-image pre-training with frozen image encoders and large language models.
\newblock In \emph{International conference on machine learning}, pages 19730--19742. PMLR, 2023{\natexlab{a}}.

\bibitem[Li et~al.(2023{\natexlab{b}})Li, He, Wang, Li, Wang, Luo, Wang, Wang, and Qiao]{videochat}
KunChang Li, Yinan He, Yi Wang, Yizhuo Li, Wenhai Wang, Ping Luo, Yali Wang, Limin Wang, and Yu Qiao.
\newblock Videochat: Chat-centric video understanding.
\newblock \emph{arXiv preprint arXiv:2305.06355}, 2023{\natexlab{b}}.

\bibitem[Li et~al.(2024{\natexlab{c}})Li, Wang, He, Li, Wang, Liu, Wang, Xu, Chen, Luo, et~al.]{li2024mvbench}
Kunchang Li, Yali Wang, Yinan He, Yizhuo Li, Yi Wang, Yi Liu, Zun Wang, Jilan Xu, Guo Chen, Ping Luo, et~al.
\newblock Mvbench: A comprehensive multi-modal video understanding benchmark.
\newblock In \emph{Proceedings of the IEEE/CVF Conference on Computer Vision and Pattern Recognition}, pages 22195--22206, 2024{\natexlab{c}}.

\bibitem[Li et~al.(2024{\natexlab{d}})Li, Wang, and Jia]{llama_vid}
Yanwei Li, Chengyao Wang, and Jiaya Jia.
\newblock Llama-vid: An image is worth 2 tokens in large language models.
\newblock In \emph{European Conference on Computer Vision}, pages 323--340. Springer, 2024{\natexlab{d}}.

\bibitem[Liang et~al.(2024)Liang, Meng, Wang, Liu, Liu, and Zhao]{liang2024end}
Jianxin Liang, Xiaojun Meng, Yueqian Wang, Chang Liu, Qun Liu, and Dongyan Zhao.
\newblock End-to-end video question answering with frame scoring mechanisms and adaptive sampling.
\newblock \emph{arXiv preprint arXiv:2407.15047}, 2024.

\bibitem[Lin et~al.(2023)Lin, Ye, Zhu, Cui, Ning, Jin, and Yuan]{video_llava}
Bin Lin, Yang Ye, Bin Zhu, Jiaxi Cui, Munan Ning, Peng Jin, and Li Yuan.
\newblock Video-llava: Learning united visual representation by alignment before projection.
\newblock \emph{arXiv preprint arXiv:2311.10122}, 2023.

\bibitem[Lin et~al.(2024)Lin, Fang, Chen, Wan, Luo, Li, Liu, and Sun]{streamingbench}
Junming Lin, Zheng Fang, Chi Chen, Zihao Wan, Fuwen Luo, Peng Li, Yang Liu, and Maosong Sun.
\newblock Streamingbench: Assessing the gap for mllms to achieve streaming video understanding.
\newblock \emph{arXiv preprint arXiv:2411.03628}, 2024.

\bibitem[Liu et~al.(2023)Liu, Li, Wu, and Lee]{llava}
Haotian Liu, Chunyuan Li, Qingyang Wu, and Yong~Jae Lee.
\newblock Visual instruction tuning.
\newblock \emph{Advances in neural information processing systems}, 36:\penalty0 34892--34916, 2023.

\bibitem[Liu et~al.(2024{\natexlab{a}})Liu, Yan, Zaharia, and Abbeel]{lwm}
Hao Liu, Wilson Yan, Matei Zaharia, and Pieter Abbeel.
\newblock World model on million-length video and language with blockwise ringattention.
\newblock \emph{arXiv preprint arXiv:2402.08268}, 2024{\natexlab{a}}.

\bibitem[Liu et~al.(2024{\natexlab{b}})Liu, Wang, Ma, Wu, Ma, Wei, Jiao, Wu, and Hu]{liu2024kangaroo}
Jiajun Liu, Yibing Wang, Hanghang Ma, Xiaoping Wu, Xiaoqi Ma, Xiaoming Wei, Jianbin Jiao, Enhua Wu, and Jie Hu.
\newblock Kangaroo: A powerful video-language model supporting long-context video input.
\newblock \emph{arXiv preprint arXiv:2408.15542}, 2024{\natexlab{b}}.

\bibitem[Liu et~al.(2025)Liu, Xie, Li, Zhao, Tang, Zheng, Liu, and Xie]{liu2025hybrid}
Zhihang Liu, Chen-Wei Xie, Pandeng Li, Liming Zhao, Longxiang Tang, Yun Zheng, Chuanbin Liu, and Hongtao Xie.
\newblock Hybrid-level instruction injection for video token compression in multi-modal large language models.
\newblock \emph{arXiv preprint arXiv:2503.16036}, 2025.

\bibitem[Luo et~al.(2023)Luo, Zhao, Yang, Dong, Li, Lu, Wang, Hu, Qiu, and Wei]{luo2023valley}
Ruipu Luo, Ziwang Zhao, Min Yang, Junwei Dong, Da Li, Pengcheng Lu, Tao Wang, Linmei Hu, Minghui Qiu, and Zhongyu Wei.
\newblock Valley: Video assistant with large language model enhanced ability.
\newblock \emph{arXiv preprint arXiv:2306.07207}, 2023.

\bibitem[Maaz et~al.(2023)Maaz, Rasheed, Khan, and Khan]{video_chatgpt}
Muhammad Maaz, Hanoona Rasheed, Salman Khan, and Fahad~Shahbaz Khan.
\newblock Video-chatgpt: Towards detailed video understanding via large vision and language models.
\newblock \emph{arXiv preprint arXiv:2306.05424}, 2023.

\bibitem[Mounir et~al.(2023)Mounir, Vijayaraghavan, and Sarkar]{streamer}
Ramy Mounir, Sujal Vijayaraghavan, and Sudeep Sarkar.
\newblock Streamer: Streaming representation learning and event segmentation in a hierarchical manner.
\newblock \emph{Advances in Neural Information Processing Systems}, 36:\penalty0 45694--45715, 2023.

\bibitem[Qian et~al.(2024)Qian, Dong, Zhang, Zang, Ding, Lin, and Wang]{qian2024streaming}
Rui Qian, Xiaoyi Dong, Pan Zhang, Yuhang Zang, Shuangrui Ding, Dahua Lin, and Jiaqi Wang.
\newblock Streaming long video understanding with large language models.
\newblock \emph{Advances in Neural Information Processing Systems}, 37:\penalty0 119336--119360, 2024.

\bibitem[Raffel et~al.(2020)Raffel, Shazeer, Roberts, Lee, Narang, Matena, Zhou, Li, and Liu]{raffel2020exploring}
Colin Raffel, Noam Shazeer, Adam Roberts, Katherine Lee, Sharan Narang, Michael Matena, Yanqi Zhou, Wei Li, and Peter~J Liu.
\newblock Exploring the limits of transfer learning with a unified text-to-text transformer.
\newblock \emph{Journal of machine learning research}, 21\penalty0 (140):\penalty0 1--67, 2020.

\bibitem[Ryoo et~al.(2023)Ryoo, Gopalakrishnan, Kahatapitiya, Xiao, Rao, Stone, Lu, Ibarz, and Arnab]{ryoo2023token}
Michael~S Ryoo, Keerthana Gopalakrishnan, Kumara Kahatapitiya, Ted Xiao, Kanishka Rao, Austin Stone, Yao Lu, Julian Ibarz, and Anurag Arnab.
\newblock Token turing machines.
\newblock In \emph{Proceedings of the IEEE/CVF Conference on Computer Vision and Pattern Recognition}, pages 19070--19081, 2023.

\bibitem[Sharma et~al.(2018)Sharma, Ding, Goodman, and Soricut]{sharma2018conceptual}
Piyush Sharma, Nan Ding, Sebastian Goodman, and Radu Soricut.
\newblock Conceptual captions: A cleaned, hypernymed, image alt-text dataset for automatic image captioning.
\newblock In \emph{Proceedings of the 56th Annual Meeting of the Association for Computational Linguistics (Volume 1: Long Papers)}, pages 2556--2565, 2018.

\bibitem[Shen et~al.(2025)Shen, Gong, He, Zhang, Liu, Zhao, and Ding]{shen2025fastvid}
Leqi Shen, Guoqiang Gong, Tao He, Yifeng Zhang, Pengzhang Liu, Sicheng Zhao, and Guiguang Ding.
\newblock Fastvid: Dynamic density pruning for fast video large language models.
\newblock \emph{arXiv preprint arXiv:2503.11187}, 2025.

\bibitem[Song et~al.(2024{\natexlab{a}})Song, Chai, Wang, Zhang, Zhou, Wu, Chi, Guo, Ye, Zhang, et~al.]{song2024moviechat}
Enxin Song, Wenhao Chai, Guanhong Wang, Yucheng Zhang, Haoyang Zhou, Feiyang Wu, Haozhe Chi, Xun Guo, Tian Ye, Yanting Zhang, et~al.
\newblock Moviechat: From dense token to sparse memory for long video understanding.
\newblock In \emph{Proceedings of the IEEE/CVF Conference on Computer Vision and Pattern Recognition}, pages 18221--18232, 2024{\natexlab{a}}.

\bibitem[Song et~al.(2024{\natexlab{b}})Song, Chai, Ye, Hwang, Li, and Wang]{song2024moviechat+}
Enxin Song, Wenhao Chai, Tian Ye, Jenq-Neng Hwang, Xi Li, and Gaoang Wang.
\newblock Moviechat+: Question-aware sparse memory for long video question answering.
\newblock \emph{arXiv preprint arXiv:2404.17176}, 2024{\natexlab{b}}.

\bibitem[Wang et~al.(2024{\natexlab{a}})Wang, Bai, Tan, Wang, Fan, Bai, Chen, Liu, Wang, Ge, et~al.]{qwen2_vl}
Peng Wang, Shuai Bai, Sinan Tan, Shijie Wang, Zhihao Fan, Jinze Bai, Keqin Chen, Xuejing Liu, Jialin Wang, Wenbin Ge, et~al.
\newblock Qwen2-vl: Enhancing vision-language model's perception of the world at any resolution.
\newblock \emph{arXiv preprint arXiv:2409.12191}, 2024{\natexlab{a}}.

\bibitem[Wang et~al.(2024{\natexlab{b}})Wang, He, Hong, Cheng, Zhang, Qi, Gu, Huang, Xu, Dong, et~al.]{Lvbench}
Weihan Wang, Zehai He, Wenyi Hong, Yean Cheng, Xiaohan Zhang, Ji Qi, Xiaotao Gu, Shiyu Huang, Bin Xu, Yuxiao Dong, et~al.
\newblock Lvbench: An extreme long video understanding benchmark.
\newblock \emph{arXiv preprint arXiv:2406.08035}, 2024{\natexlab{b}}.

\bibitem[Wang et~al.(2023)Wang, Zhang, Wang, Bhattacharya, Fu, and Wu]{wang2023vaquita}
Yizhou Wang, Ruiyi Zhang, Haoliang Wang, Uttaran Bhattacharya, Yun Fu, and Gang Wu.
\newblock Vaquita: Enhancing alignment in llm-assisted video understanding.
\newblock \emph{arXiv preprint arXiv:2312.02310}, 2023.

\bibitem[Wang et~al.(2024{\natexlab{c}})Wang, Xie, Liu, and Zheng]{wang2024videollamb}
Yuxuan Wang, Cihang Xie, Yang Liu, and Zilong Zheng.
\newblock Videollamb: Long-context video understanding with recurrent memory bridges.
\newblock \emph{arXiv preprint arXiv:2409.01071}, 2024{\natexlab{c}}.

\bibitem[Xu et~al.(2016)Xu, Mei, Yao, and Rui]{msrvtt}
Jun Xu, Tao Mei, Ting Yao, and Yong Rui.
\newblock Msr-vtt: A large video description dataset for bridging video and language.
\newblock In \emph{Proceedings of the IEEE conference on computer vision and pattern recognition}, pages 5288--5296, 2016.

\bibitem[Xu et~al.(2024)Xu, Zhao, Zhou, Lin, Ng, and Feng]{pllava}
Lin Xu, Yilin Zhao, Daquan Zhou, Zhijie Lin, See~Kiong Ng, and Jiashi Feng.
\newblock Pllava: Parameter-free llava extension from images to videos for video dense captioning.
\newblock \emph{arXiv preprint arXiv:2404.16994}, 2024.

\bibitem[Yang et~al.(2024)Yang, Yang, Zhang, Hui, Zheng, Yu, Li, Liu, Huang, Wei, et~al.]{yang2024qwen2}
An Yang, Baosong Yang, Beichen Zhang, Binyuan Hui, Bo Zheng, Bowen Yu, Chengyuan Li, Dayiheng Liu, Fei Huang, Haoran Wei, et~al.
\newblock Qwen2. 5 technical report.
\newblock \emph{arXiv preprint arXiv:2412.15115}, 2024.

\bibitem[Yao et~al.(2024)Yao, Yu, Zhang, Wang, Cui, Zhu, Cai, Li, Zhao, He, et~al.]{Minicpm}
Yuan Yao, Tianyu Yu, Ao Zhang, Chongyi Wang, Junbo Cui, Hongji Zhu, Tianchi Cai, Haoyu Li, Weilin Zhao, Zhihui He, et~al.
\newblock Minicpm-v: A gpt-4v level mllm on your phone.
\newblock \emph{arXiv preprint arXiv:2408.01800}, 2024.

\bibitem[Ye et~al.(2023)Ye, Xu, Xu, Ye, Yan, Zhou, Wang, Hu, Shi, Shi, et~al.]{mplug_o1}
Qinghao Ye, Haiyang Xu, Guohai Xu, Jiabo Ye, Ming Yan, Yiyang Zhou, Junyang Wang, Anwen Hu, Pengcheng Shi, Yaya Shi, et~al.
\newblock mplug-owl: Modularization empowers large language models with multimodality.
\newblock \emph{arXiv preprint arXiv:2304.14178}, 2023.

\bibitem[Ye et~al.(2024)Ye, Xu, Ye, Yan, Hu, Liu, Qian, Zhang, and Huang]{mplug_o2}
Qinghao Ye, Haiyang Xu, Jiabo Ye, Ming Yan, Anwen Hu, Haowei Liu, Qi Qian, Ji Zhang, and Fei Huang.
\newblock mplug-owl2: Revolutionizing multi-modal large language model with modality collaboration.
\newblock In \emph{Proceedings of the ieee/cvf conference on computer vision and pattern recognition}, pages 13040--13051, 2024.

\bibitem[Zacks et~al.(2007)Zacks, Speer, Swallow, Braver, and Reynolds]{est}
Jeffrey~M Zacks, Nicole~K Speer, Khena~M Swallow, Todd~S Braver, and Jeremy~R Reynolds.
\newblock Event perception: a mind-brain perspective.
\newblock \emph{Psychological bulletin}, 133\penalty0 (2):\penalty0 273, 2007.

\bibitem[Zhang et~al.(2023)Zhang, Li, and Bing]{video_llama}
Hang Zhang, Xin Li, and Lidong Bing.
\newblock Video-llama: An instruction-tuned audio-visual language model for video understanding.
\newblock \emph{arXiv preprint arXiv:2306.02858}, 2023.

\bibitem[Zhang et~al.(2024{\natexlab{a}})Zhang, Wang, Tang, Liu, Feng, Dai, and Jin]{flash_vstream}
Haoji Zhang, Yiqin Wang, Yansong Tang, Yong Liu, Jiashi Feng, Jifeng Dai, and Xiaojie Jin.
\newblock Flash-vstream: Memory-based real-time understanding for long video streams.
\newblock \emph{arXiv preprint arXiv:2406.08085}, 2024{\natexlab{a}}.

\bibitem[Zhang et~al.(2024{\natexlab{b}})Zhang, Zhang, Li, Zeng, Yang, Zhang, Wang, Tan, Li, and Liu]{longva}
Peiyuan Zhang, Kaichen Zhang, Bo Li, Guangtao Zeng, Jingkang Yang, Yuanhan Zhang, Ziyue Wang, Haoran Tan, Chunyuan Li, and Ziwei Liu.
\newblock Long context transfer from language to vision.
\newblock \emph{arXiv preprint arXiv:2406.16852}, 2024{\natexlab{b}}.

\bibitem[Zhang et~al.(2024{\natexlab{c}})Zhang, Li, Liu, Lee, Gui, Fu, Feng, Liu, and Li]{llava_next}
Yuanhan Zhang, Bo Li, haotian Liu, Yong~jae Lee, Liangke Gui, Di Fu, Jiashi Feng, Ziwei Liu, and Chunyuan Li.
\newblock Llava-next: A strong zero-shot video understanding model, 2024{\natexlab{c}}.

\bibitem[Zhou et~al.(2024)Zhou, Shu, Zhao, Wu, Xiao, Yang, Xiong, Zhang, Huang, and Liu]{mlvu}
Junjie Zhou, Yan Shu, Bo Zhao, Boya Wu, Shitao Xiao, Xi Yang, Yongping Xiong, Bo Zhang, Tiejun Huang, and Zheng Liu.
\newblock Mlvu: A comprehensive benchmark for multi-task long video understanding.
\newblock \emph{arXiv preprint arXiv:2406.04264}, 2024.

\bibitem[Zhu et~al.(2023)Zhu, Chen, Shen, Li, and Elhoseiny]{minigpt}
Deyao Zhu, Jun Chen, Xiaoqian Shen, Xiang Li, and Mohamed Elhoseiny.
\newblock Minigpt-4: Enhancing vision-language understanding with advanced large language models.
\newblock \emph{arXiv preprint arXiv:2304.10592}, 2023.

\bibitem[Zhu et~al.(2025)Zhu, Wang, Chen, Liu, Ye, Gu, Duan, Tian, Su, Shao, et~al.]{internvl3}
Jinguo Zhu, Weiyun Wang, Zhe Chen, Zhaoyang Liu, Shenglong Ye, Lixin Gu, Yuchen Duan, Hao Tian, Weijie Su, Jie Shao, et~al.
\newblock Internvl3: Exploring advanced training and test-time recipes for open-source multimodal models.
\newblock \emph{arXiv preprint arXiv:2504.10479}, 2025.

\end{thebibliography}
}

% WARNING: do not forget to delete the supplementary pages from your submission 
% \input{sec/X_suppl}

\end{document}